\documentclass{article} 

\usepackage{aaai2026}  
\usepackage{times}  
\usepackage{helvet}  
\usepackage{courier}  
\usepackage[hyphens]{url}  
\usepackage{graphicx} 
\usepackage{natbib}  
\usepackage{caption} 
\usepackage{algorithm}
\usepackage{algorithmic}
\usepackage{url}
\usepackage{amssymb}
\usepackage{amsmath}
\usepackage{graphicx}
\usepackage{tabularx}
\usepackage{soul}
\usepackage[dvipsnames]{xcolor}
\definecolor{globalcolor}{RGB}{0, 176, 80}
\definecolor{domaincolor}{RGB}{0, 112, 192}
\usepackage{multirow}
\usepackage{booktabs}

\usepackage{newfloat}
\usepackage{listings}
\DeclareCaptionStyle{ruled}{labelfont=normalfont,labelsep=colon,strut=off} 
\floatstyle{ruled}
\newfloat{listing}{tb}{lst}{}
\floatname{listing}{Listing}
\title{XAgents: A Unified Framework for Multi-Agent Cooperation via IF-THEN Rules and Multipolar Task Processing Graph}
\author{
    Hailong Yang\textsuperscript{\rm 1}\equalcontrib,Mingxian Gu\textsuperscript{\rm 1}\equalcontrib, Jianqi Wang\textsuperscript{\rm 1}, Guanjin Wang\textsuperscript{\rm 2}, Zhaohong Deng\textsuperscript{\rm 1}\thanks{Corresponding author.}\\
}
\affiliations{
    \textsuperscript{\rm 1} School of Artificial Intelligence and Computer Science, 
     Jiangnan University, Wuxi 214122, China\\
    \textsuperscript{\rm 2} School of Information Technology, 
  Murdoch University, WA, 6150, Australia \\
    \{yanghailong, gumingxian, wangjianqi\}@stu.jiangnan.edu.cn, Guanjin.Wang@murdoch.edu.au, dengzhaohong@jiangnan.edu.cn\\
}

\usepackage{bibentry}

\begin{document}

\maketitle

\begin{abstract}
The rapid advancement of Large Language Models (LLMs) has significantly enhanced the capabilities of Multi-Agent Systems (MAS) in supporting humans with complex, real-world tasks. However, MAS still face challenges in effective task planning when handling highly complex tasks with uncertainty, often resulting in misleading or incorrect outputs that hinder task execution. To address this, we propose XAgents, a unified multi-agent cooperative framework built on a multipolar task processing graph and IF-THEN rules. 
XAgents uses the multipolar task processing graph to enable dynamic task planning and handle task uncertainty. 
During subtask processing, it integrates domain-specific IF-THEN rules to constrain agent behaviors, while global rules enhance inter-agent collaboration. We evaluate the performance of XAgents across three distinct datasets, demonstrating that it consistently surpasses state-of-the-art single-agent and multi-agent approaches in both knowledge-typed and logic-typed question-answering tasks. The codes for XAgents are available at: \url{https://github.com/AGI-FHBC/XAgents}.
\end{abstract}


\section{Introduction}

With the rapid advancement of Large Language Models (LLMs), the LLM-based Multi-Agent System (MAS) \cite{li_survey_2024} has emerged as a promising paradigm for tackling complex problems through interaction and cooperation among multiple autonomous agents. 
MAS improves decision-making and task execution ability by leveraging human-like reasoning patterns and collaborative dynamics. As MAS research progresses, role-based structures have been introduced to optimize task processing \cite{lan_stance_2024}. In such systems, a designated planning agent is responsible for decomposing tasks and formulating a task graph, thereby improving the overall execution efficiency of the MAS.

Although MAS offers significant advantages, it still faces key challenges:
\begin{itemize}
    \item MAS often struggles with planning when dealing with complex tasks that involve uncertainty, leading to inefficient or incomplete task execution.
    \item Due to LLM hallucination, agents may produce misleading or incorrect outputs, compromising the overall performance of the system.
\end{itemize}

\begin{figure}[t]
  \includegraphics[width=\columnwidth]{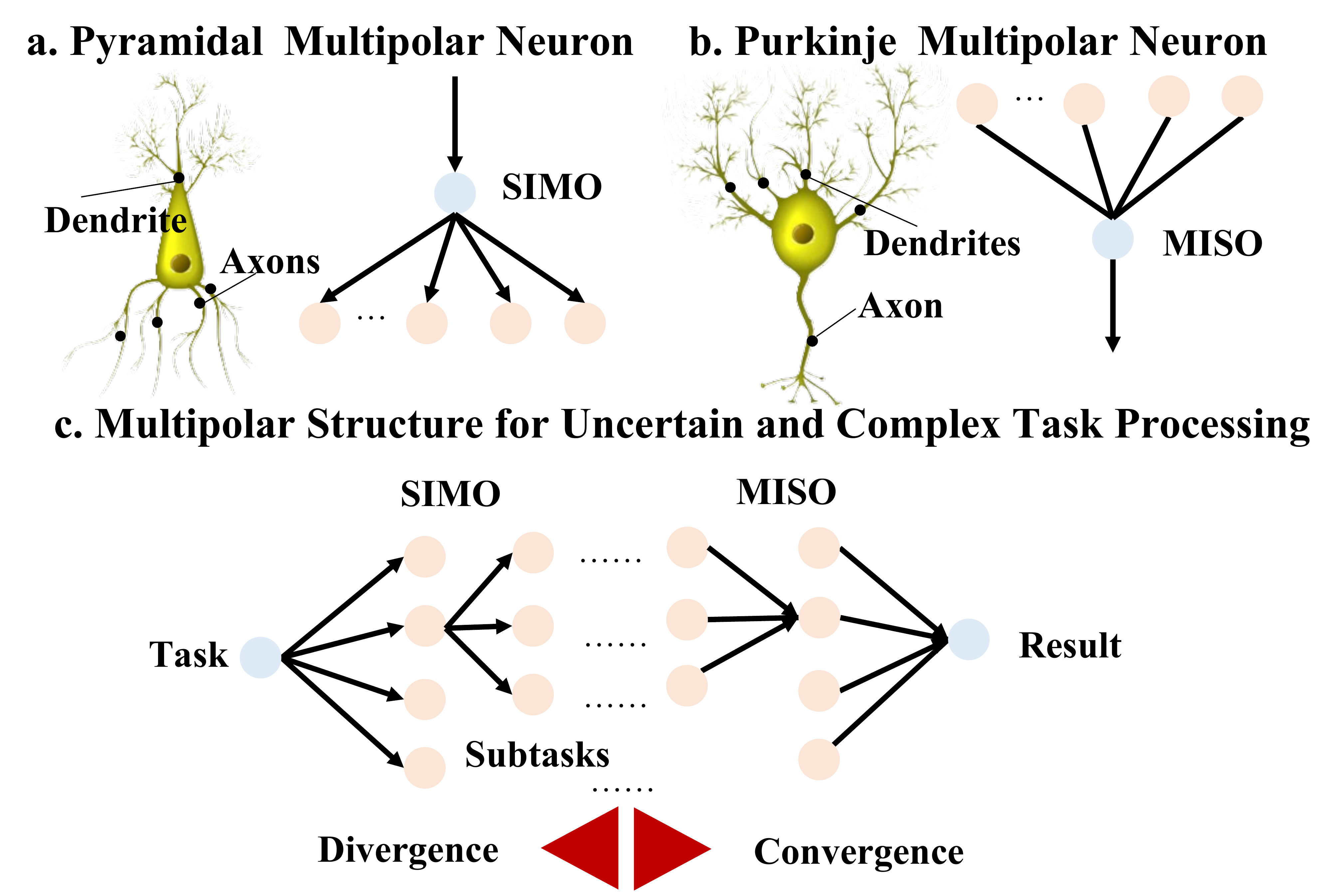}
  \caption{Multipolar structure for uncertain task processing,
the abstract structures of multipolar neurons include (a) Single Input Multiple Output (SIMO) and (b) Multiple Input Single Output (MISO), (c) presents a multipolar divergence-convergence structure designed for handling uncertain tasks.}
  \label{fig:multipolar}
\end{figure}

To address these challenges, we draw inspiration from the biological structure of the human brain, which offers a natural model for handling uncertainty and distributed processing. Specifically, the abstract Single Input Multiple Output (SIMO) and Multiple Input Single Output (MISO) structures inspired by Multipolar Neurons (MNs), such as pyramidal neurons \cite{jossin_reelin_2011}(Fig. \ref{fig:multipolar} a) and Purkinje neurons \cite{herndon_fine_1963} (Fig. \ref{fig:multipolar} b) model diverge-converge information flows. The structures enable robust signal transmission under uncertainty.

Inspired by MNs, we apply a SIMO and MISO-based multipolar task structure in MAS. When dealing with uncertain tasks, SIMO can enable divergent thinking \cite{liang_encouraging_2024} and explore multiple potential task paths concurrently, increasing the likelihood of successful task execution, as shown in Fig. \ref{fig:multipolar} c.

Building on the biologically inspired multipolar structure for managing uncertainty, we further consider incorporating IF-THEN rules to introduce domain-specific constraints and guide agent behavior. The structure of IF-THEN rules is a conditional statement composed of an antecedent (IF) and a consequent (THEN), commonly used to direct system behavior \cite{liu_rule_2017}. In constrained environments, these rules encode domain-specific knowledge and constraints to ensure that agent decisions are consistent, interpretable, and aligned with overall task goals. 
By embedding semantic reasoning based on multiple domains, IF-THEN rules enhance the accuracy of outputs and effectively mitigate hallucination-induced errors from LLMs.

In summary, to address the two aforementioned challenges—task planning under uncertainty and the lack of constraints for LLM hallucination- we propose XAgents, a unified multi-agent collaborative framework that integrates the multipolar task structure with rule-based decision-making. XAgents consists of two key components: (i) a multipolar task processing graph, inspired by biological signal pathways (SIMO and MISO), and (ii) a rule-based mechanism that incorporates domain-specific IF-THEN rules to guide agent behavior.

This paper contributes by
\begin{itemize}
    \item proposing a multipolar structured task processing graph based on SIMO and MOSI, offering a novel approach for handling uncertain tasks.
    \item introducing IF-THEN rules into MAS to constrain and guide agent behavior, further improving decision quality and reducing hallucinations.
    \item presenting XAgents, a unified multi-agent cooperative framework that integrates a multipolar structure with rule-based reasoning. 
\end{itemize}


\section{Related Work}
\label{sec:related}

 
The research on agents contains two key fields, \emph{single agent} and \emph { multiple agents}. A single agent focuses on reasoning mechanisms. The notable works are Chain-of-Thought (CoT) \cite{wei_chain--thought_2022}, least-to-most prompting \cite{zhou_least--most_2023}, zero-sample CoT \cite{kojima_large_2022}, self-consistent reasoning mechanisms \cite{wang_self-consistency_2023}, and iterative self-refine by feedback \cite{Madaan_self_2023}. Multiple agents focus on multi-agent frameworks with cooperative and collaborative mechanisms \cite{zhang_proagent_2024}. SPP \cite{wang_unleashing_2024} is a multi-persona mechanism to process complex tasks in multi-turn self-collaboration. AutoAgents \cite{chen_autoagents_2024} draws a connection between tasks and roles by dynamically generating multiple agents.TDAG \cite{wang_tdag_2025} focuses on the dynamic decomposition of complex problems and the execution of sub-tasks, while AgentNet \cite{yang_agentnet_2025} enables efficient collaboration in multi-agent systems through a decentralized coordination mechanism.


Rule-based systems have been used to capture and refine human expertise \cite{davis_origin_1984,hayes-roth_rule-based_1985}. The systems have been further developed in the field of Mixture of Experts (MoE) systems \cite{jacobs_adaptive_1991,yuksel_twenty_2012}. 
 Wang et al. proposed the Bayesian Rule Set (BRS) based on Bayesian theory and proved its interpretability \cite{wang_bayesian_2017}. Liu et al. investigated the interpretable representations of rule-based networks to discover deep knowledge \cite{liu_rule_2017}. RuleXAI \cite{macha_rulexaipackage_2022} is a rule-based modeling interpretable approach.

\section{XAgents: A Multipolar Rule-Based Multi-Agent Cooperative Framework}
\label{sec:xagents}
XAgents is a multi-agent cooperative framework that integrates two key modules: the Multipolar Task Processing Graph (MTPG) and the IF-THEN Rule-based Decision Mechanism (ITRDM). 


\begin{figure*}[t]
  \centerline{\includegraphics[width=1.0\linewidth]{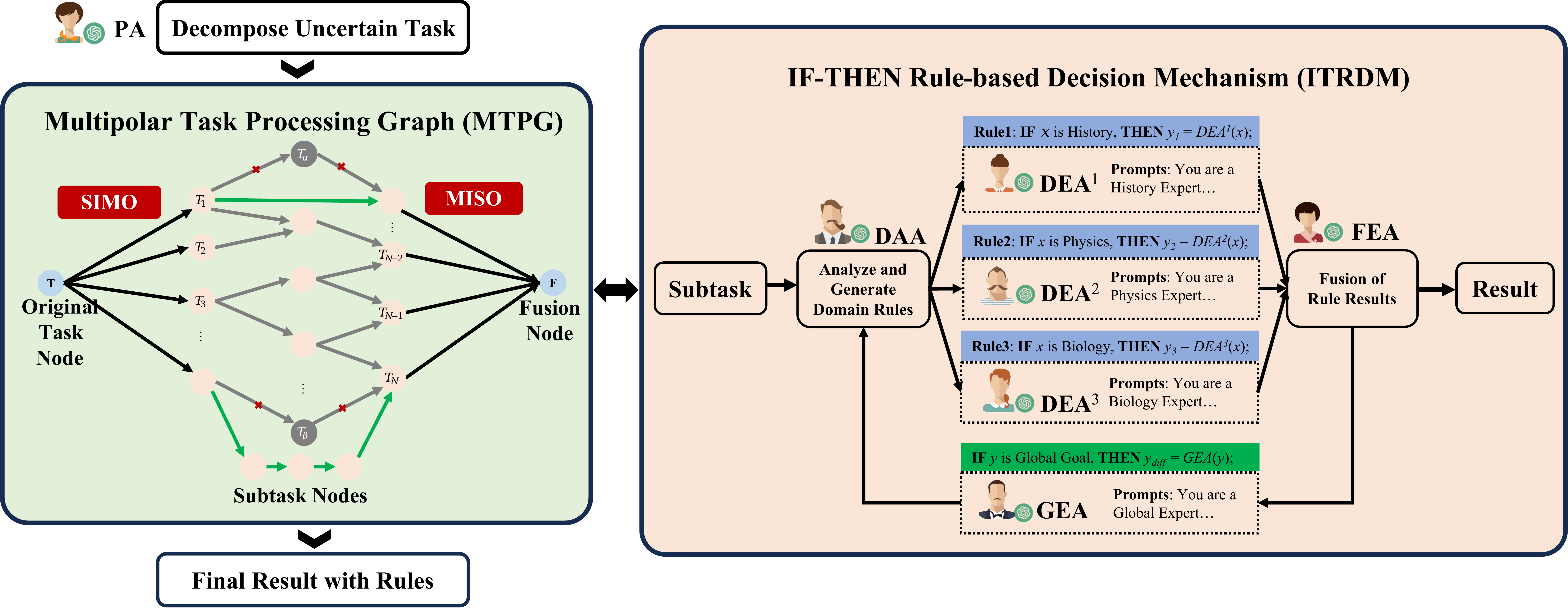}}
  \caption{Structure of XAgents. which incorporated two key components: Multipolar Task-Processing Graph (MTPG, Section \ref{subsec:mtpg}) and IF-THEN Rule-based Decision Mechanism (ITRDM, Section \ref{subsec:itrdm}). XAgents includes Planner Agent (PA), Domain Analyst Agent (DAA), and Domain
Expert Agents (DEA), Fusion Expert Agent (FEA), and
Global Expert Agent (GEA).}
  \label{fig:xagents}
\end{figure*}

\subsection{Multipolar Task Processing Graph (MTPG)}
\label{subsec:mtpg}
To tackle uncertain and complex tasks, we developed the MTPG derived from SIMO and MISO. MTPG is a directed acyclic graph with unweighted edges that encode dependency only. MTPG incorporates three types of nodes: an original task node, subtask nodes, and a fusion node. SIMO enables divergent decomposition of tasks, while MISO facilitates convergent fusion of subtask results. The system first constructs a divergent graph structure through hierarchical SIMO, and then forms a convergent graph structure via hierarchical MISO, thereby completing a closed-loop process from task decomposition to result integration.

In the planning phase, Planner Agent (PA) constructs an original MTPG, and the relationships of subtasks can be expressed as the edges of a graph. Since the multipolar structure can balance the inputs and outputs of nodes, MTPG uses multi-input balancing to mitigate task-level uncertainty. 

The original task node represents the original uncertain task node $T$ at the root of MTPG, linked by simple subtask nodes. Subtask nodes $\{T_{1},…, T_{N}\}$ constitute the primary part of MTPG, and the final result is obtained by fusing the outcomes of adjacent subtask nodes linked by the fusion node $F$, as shown in Eq. (\ref{eq:graph}).

\begin{align}
    \mathcal{V}&=\{T,T_{1},…,T_{N},F\}\\
    \mathcal{E}&=\{(T,T_{1}),(T,T_{2}),...,(T_{N},F)\}\\
    \mathcal{G}&=PA(x) \label{eq:graph}
\end{align}
where $x$ is the input uncertain task, $\mathcal{G}=(\mathcal{V},\mathcal{E})$ is MTPG, $PA(.)$ is PA processing unit function for planning tasks, $\mathcal{V}$ is the node set of MTPG, $\mathcal{E}$ is the edges set of MTPG.

\subsection{IF-THEN Rule-based Decision Mechanism (ITRDM)}
\label{subsec:itrdm}


Each subtask node is addressed through a combination of IF-THEN rules and collaborative efforts among agents. The agents include the Domain Analyst Agent (DAA), multiple Domain Expert Agents (DEA), the Fusion Expert Agent (FEA) and the Global Expert Agent (GEA). 

\subsubsection{IF-THEN Rule Construction}
\label{subsubsec:if then rule cons}
An IF-THEN rule is composed of two parts: the antecedent (IF-part) and the consequent (THEN-part). The antecedent determines the activation strength of the rule and the consequent determines the resulting output or conclusion \cite{sun_robust_1995}.

In XAgents, the execution of subtasks is driven by IF-THEN rules. Each subtask comprises multiple domain rules and a global rule. 
As illustrated in Eq. (\ref{eq:rule}), the IF-parts of the domain rules are expressed in natural language constructed using domain-specific terminology. These IF-parts are used to compute the membership degree of the input concerning the given domain \cite{dombi_membership_1990}. The corresponding THEN parts guide the DEAs in producing subtask outcomes based on domain knowledge. The IF-parts and THEN-parts of the domain rules are automatically generated by the DAA.

The outputs of the domain rules are constrained by the global rule, which follows a similar IF-THEN structure. The IF-part of the global rule computes the membership degree of the fused result concerning the global goal, as defined by the PA. The THEN-part outputs the differences between the fused result and the global goal. An illustrative example is shown below:

\begin{equation}
\begin{split}
    &\textbf{\textcolor{domaincolor}{IF}}\, x_{\scriptscriptstyle T_i}\, is\, Domain^{1},\, \textbf{\textcolor{domaincolor}{THEN}} \,
    y_{\scriptscriptstyle T_i}^{1}=DEA^{1}(x_{\scriptscriptstyle T_i},\mathcal{P}_{\scriptscriptstyle T_i}); \\
    &\textbf{\textcolor{domaincolor}{IF}}\, x_{\scriptscriptstyle T_i}\, is\, Domain^{2},\, \textbf{\textcolor{domaincolor}{THEN}}\,
    y_{\scriptscriptstyle T_i}^{2}=DEA^{2}(x_{\scriptscriptstyle T_i},\mathcal{P}_{\scriptscriptstyle T_i}); \\
    &\textbf{\textcolor{domaincolor}{IF}}\, x_{\scriptscriptstyle T_i}\, is\, Domain^{3}  ,\,\textbf{\textcolor{domaincolor}{THEN}} \,
    y_{\scriptscriptstyle T_i}^{3}=DEA^{3}(x_{\scriptscriptstyle T_i},\mathcal{P}_{\scriptscriptstyle T_i}); \\
    &\textbf{\textcolor{globalcolor}{IF}}\, y_{\scriptscriptstyle T_i}\,  is\, Global Goal,\,\textbf{\textcolor{globalcolor}{THEN}} \,
    y_{\scriptscriptstyle diff}=GEA(y_{\scriptscriptstyle T_i}); 
    \label{eq:rule}
\end{split}
\end{equation}
where $is$ denotes the operator used to compute membership degree, $x_{\scriptscriptstyle T_i}$ is the input for $i$-th subtask Node, $i=[1,…,K]$, $K$ is the total of domain rules in the subtask, $\mathcal{P}_{\scriptscriptstyle T_i}$ is the result set of the previous nodes, $y_{\scriptscriptstyle T_i}$ is the fused result of $K$ rules.

The DEAs for subtasks are integrated into IF-THEN rule-based systems. In the planning phase, PA does not generate detailed domain rules. Instead, it delegates the task of generating domain rules to the DAA, following the domain analysis for rule antecedents and initialization of DEAs’ prompts for the rule consequent, as shown in Eq. (\ref{eq:daa}). Consequently, during the executing phase, each subtask node is distant and dynamic, as shown in Eq. (\ref{eq:run}). FEA fuses the domain responses from DEAs with different domain-specific knowledge, as shown in Eq. (\ref{eq:fea}).
\begin{align}
    rules_{\scriptscriptstyle T_{i}}&=DAA(x_{\scriptscriptstyle T_i}) \label{eq:daa}\\
    \boldsymbol{\lambda}_{T_i}&=Run(rules_{\scriptscriptstyle T_i}\vert x_{\scriptscriptstyle T_i},\mathcal{P}_{\scriptscriptstyle T_i}) \label{eq:run}\\
    y_{\scriptscriptstyle T_i}&=FEA_{sub}(\boldsymbol{\lambda}_{\scriptscriptstyle T_i}) \label{eq:fea}
\end{align}
where $DAA(.)$ is the DAA processing unit, $FEA_{sub}(.)$ is the FEA processing unit for rules , $rules_{\scriptscriptstyle T_{i}}=\{r_{ \scriptscriptstyle T_i}^{1},r_{\scriptscriptstyle T_i}^{2},...,r_{\scriptscriptstyle T_i}^{K}\}$ is the domain rule set of the $i$-th subtask defined in Eq. (\ref{eq:rule}), $Run(.)$ is the rule executing function, $\boldsymbol{\lambda}_{\scriptscriptstyle T_i}$ is the result set of the rules, $y_{\scriptscriptstyle T_{i}}$ is the fused result of the $i$-th subtask.

At the end of the executing phase, FEA fuses the outcomes of the previous subtask nodes to output the final result, as illustrated in Eq. (\ref{eq:fea2}).
\begin{equation}
    y_{\scriptscriptstyle F}=FEA_{final}(\mathcal{P}_{\scriptscriptstyle F}) \label{eq:fea2}
\end{equation}
where $FEA_{final}$ is the FEA processing unit of the fusion node, $\mathcal{P}_{\scriptscriptstyle F}$ is the input set for the fusion node, and $y_{\scriptscriptstyle F}$ is the final result of MTPG. 


\subsubsection{Rule-based Decision-making}
\label{subsubsec:logical reasoning}
Due to the hallucination problem inherent in LLMs, LLM-based agents may produce inaccurate outputs or generate content misaligned with the intended task objectives. To mitigate this, in XAgents, the IF-THEN rules restrict the behaviors of the agents. 
The IF parts of domain rules are concerned with the calculation of the domain membership of a subtask, while the THEN parts incorporate DEAs to tackle the domain-specific tasks. 

In contrast to traditional numerical methods, membership is quantized into discrete term-based labels \emph{\{High (H), Sub-High (SH), Medium (M), Mid-Low (ML), Lower (Lr), Low (L)\}} via LLM-based reasoning.

When semantic conflicts arise among rule outputs, XAgents employs a two-layer mechanism: (i) the semantic with more votes is retained; (ii) the semantic with a higher membership degree is retained. Through mutual verification and semantic confrontation of these results, FEA reduces uncertainties and ambiguities of the fused result. Thus, through the execution and fusion of domain rules, XAgents performs logical reasoning and generates reliable and contextually grounded outputs, which also limits hallucinations in LLMs. XAgents decide whether to reprocess a task based on the membership degree of the global rules. The rule-based logical reasoning and decision-making process addresses the ambiguity and uncertainty inherent in the task.

\subsubsection{Global Goal Alignment}
\label{subsubsec:global goal align}

XAgents decomposes complex tasks into simpler subtasks and coordinates the subtasks through global rules. The global goal is defined by PA based on the content of the original task, as shown in Eq. (\ref{eq:gea}). When the membership degree of the global rule falls below the threshold \emph{ML}, the global rule prompts the system to revisit the subtask. The output of the global rule is combined with the subtask for reprocessing, as shown in Eq. (\ref{eq:daa2}). If the membership degree of the subtask exceeds the threshold, the new $ y_{\scriptscriptstyle T_i}$ is passed as the node output to the next subtask node.
\begin{align}
    y_{\scriptscriptstyle diff}&=Run(rule_{glb}\vert y_{\scriptscriptstyle T_i}) \label{eq:gea}\\
    rules_{\scriptscriptstyle T_i}&=DAA(Concat(x_{\scriptscriptstyle T_i},y_{\scriptscriptstyle diff})) \label{eq:daa2}
\end{align}
where $rule_{glb}$ is the last IF-THEN rule in Eq. (\ref{eq:rule}).
\subsubsection{Autonomous Path Reconstruction}
\label{subsubsec:atuo path rec}
MTPG is initialized by PA. However, due to the complexity and uncertainty of the origin task, PA cannot guarantee that the subtasks and their relationships within MTPG are optimal. Thus, XAgents dynamically restructure the task processing paths only when the membership degree according to the global rule falls below \emph{Mid-Low (ML)}. 

When the output of a subtask node contains contradictions or ambiguities, XAgents utilizes multi-domain-rule reasoning within the node to reduce incorrect or ambiguous content. 
If a subtask node still fails to align with the global goal after several reprocessing steps, XAgents will dynamically adjust the local structure of the MTPG to maintain consistent task execution. 

Node failure can arise under two primary scenarios: (i) the subtask is irrelevant to the global goal and should be removed, e.g., node $T_{\alpha}$ at the top of the light-green box in Fig. \ref{fig:xagents}; 
(ii) the subtask remains too uncertain or complex to yield a valid result, e.g., node $T_{\beta}$ in Fig. \ref{fig:xagents}.
In the latter case, node $T_{\beta}$ should be decomposed into several simpler subtask nodes, as shown in Eq. (\ref{eq:PA2}). These simpler subtask nodes constitute a new task path, as shown in Eq. (\ref{eq:g update}). 
\begin{align}
    \mathcal{G}_{\beta}&=PA(x_{\scriptscriptstyle T_{\beta}}) \label{eq:PA2} \\
    \mathcal{G}^{\prime}&=\mathcal{G}+\mathcal{G}_{\beta} \label{eq:g update}
\end{align}
where $G_\beta=(\mathcal{V}_{\beta}, \mathcal{E}_{\beta})$ is task processing graph for the task $T_\beta$, $\mathcal{V}_{\beta}=\{T_{\beta_1},…,T_{\beta_{\scriptscriptstyle M}}\}$, $\mathcal{E}_\beta=\{(T_{\beta_1},T_{\beta_2}),…,(T_{\beta_{{\scriptscriptstyle M-1}}},T_{\beta_{\scriptscriptstyle M}})\}$, $M$ is the total of subtasks of the task $T_\beta$, $\mathcal{G}^{\prime}$ is the adjusted MTPG for the task $T$.

Overall, ITRDM is a rule-based mechanism designed to guide and constrain agent behaviors through explicit rule-based reasoning, to mitigate misleading or incorrect outputs. 

\section{Experiment}
\label{sec:expriment}

\subsection{Preliminary}
\subsubsection{Datasets}
we conduct the experiments on three question-answering datasets of \cite{wang_unleashing_2024} including a knowledge-typed dataset (\emph{Trivia Creative Writing, TCW}), a knowledge-logic-mixed dataset (\emph{Codenames Collaborative, CC}), and a logic-typed dataset (\emph{Logic Grid Puzzle, LGP}). In TCW, there are two subsets (TCW5 and TCW10), $N=5$ and $N=10$, respectively,  and $N$ is the total number of trivia questions for each task sample. TCW5/10: 100 each, LGP: 200, CC: 50 task samples. The details are shown in \emph{Appendix A}. 

\subsubsection{Metric}
Drawing on the approach of \cite{wang_unleashing_2024}, we conduct string matching with the veridical target answers for each question on the generated output. The generalized form is shown as Eq. (\ref{eq:score}).
\begin{equation}
    Score=\dfrac{A_{correct}}{N_q} \label{eq:score}
\end{equation}
where $N_q$ is the number of questions, $A_{correct}$ is the number of correct answer mentions, $Score$ is the metrics score for the tasks.

\subsubsection{Baselines}
We compare our approach with three single-agent methods, including Standard (single-turn QA), CoT, and Self-Refine, as well as four multi-agent methods, including SPP, AutoAgents, TDAG, and AgentNet.

\subsubsection{Settings}
The default LLM for Agents is GPT-4. The experiments were performed on a server equipped with 2 * Intel 8352V CPU (2.1 GHz, 36 cores), 256 GB RAM. We chose 20 domains for rule generation. The domain names and LLM temperature are shown in \emph{Appendix B}.

\subsection{Performance Comparison Analysis}

\begin{table}[h]\
  \setlength{\tabcolsep}{2mm}
  \centering
  \begin{tabular}{lcccc}
    \toprule[2pt]
    \textbf{Methods} &\textbf{TCW5} &\textbf{TCW10} &\textbf{CC} &\textbf{LGP}\\
    \midrule[1pt]
    \textbf{Single-Agent}\\
    \midrule[0.5pt]
    Standard&	74.6&	77.0&    75.4&   57.7\\
    CoT \shortcite{wei_chain--thought_2022}&	    67.1&	68.5&    72.7&   65.8\\
    Self-Refine \shortcite{Madaan_self_2023}& 73.9&	76.9&    75.3&   60.0\\
    \midrule[0.5pt]
    \textbf{Multi-Agent}\\
    \midrule[0.5pt]
    SPP \shortcite{wang_unleashing_2024}&	    79.9&	84.7&    79.0&   68.3\\
    AutoAgents \shortcite{chen_autoagents_2024}&	82.0&	85.3&    81.4&   71.8\\
    TDAG \shortcite{wang_tdag_2025}&	    78.4&	80.7&    75.9&   67.0\\
    AgentNet \shortcite{yang_agentnet_2025}&	82.1&	86.1&    82.3&   72.1\\
    XAgents (ours)&	   \textbf{84.4}&	\textbf{88.1}& \textbf{83.5}&  \textbf{75.0}\\
    \bottomrule[2pt]
  \end{tabular}
  \caption{\label{tab:per comp} Performance comparison using GPT-4.}
\end{table}

To evaluate the overall performance of XAgents, we designed a series of experiments based on question-answering task datasets. Each dataset focuses on different aspects of agent capabilities. TCW5 and TCW10 primarily test the method's ability to answer knowledge-based questions. LGP is designed to assess logical reasoning ability, while CC evaluates both knowledge comprehension and logical reasoning simultaneously.

As shown in Table \ref{tab:per comp}, XAgents consistently outperforms both single-agent and multi-agent baselines across knowledge-based and reasoning-based datasets. \emph{Appendix C} presents the details of task processing on XAgents. From the experimental results, we observe the following:

\begin{itemize}
    \item XAgents demonstrates better performance on knowledge-based questions compared to reasoning-based ones. The superior performance of XAgents is attributed to its use of MTPG to deconstruct complex questions and its application of domain rules to elicit latent knowledge from the LLM, thereby enabling accurate responses through multiple domain-specific knowledge.
    \item On logic-based question-answering tasks, XAgents outperforms the state-of-the-art AgentNet. This improvement can be attributed to the use of global IF-THEN rules, which guide the agents to generate responses aligned with the global goal. This reduces misleading or inconsistent logic to enhance the logical reasoning capabilities of XAgents.
    \item The performance gains of XAgents are more significant on tasks involving both knowledge and logic than on tasks that involve only one of these aspects. This underscores its synergy between domain knowledge and logical inference. 
\end{itemize}

Notably, CoT performs worse than Standard on TCW5/10, potentially due to over-reasoning that hinders accuracy in knowledge-based questions. In contrast, for logic-based tasks such as LGP, CoT outperforms Standard.

In summary, the integration of IF-THEN rules and MTPG allows XAgents to outperform the baselines in both knowledge-intensive and reasoning-intensive tasks. 


 \emph{Appendix D and E} provide a structural comparison between XAgents and other multi-agent methods, as well as a performance comparison on GPT-3.5 and Llama 3.1. The results further demonstrate the superiority of XAgents.







\begin{figure*}[t]
  \centerline{\includegraphics[width=0.8\linewidth]{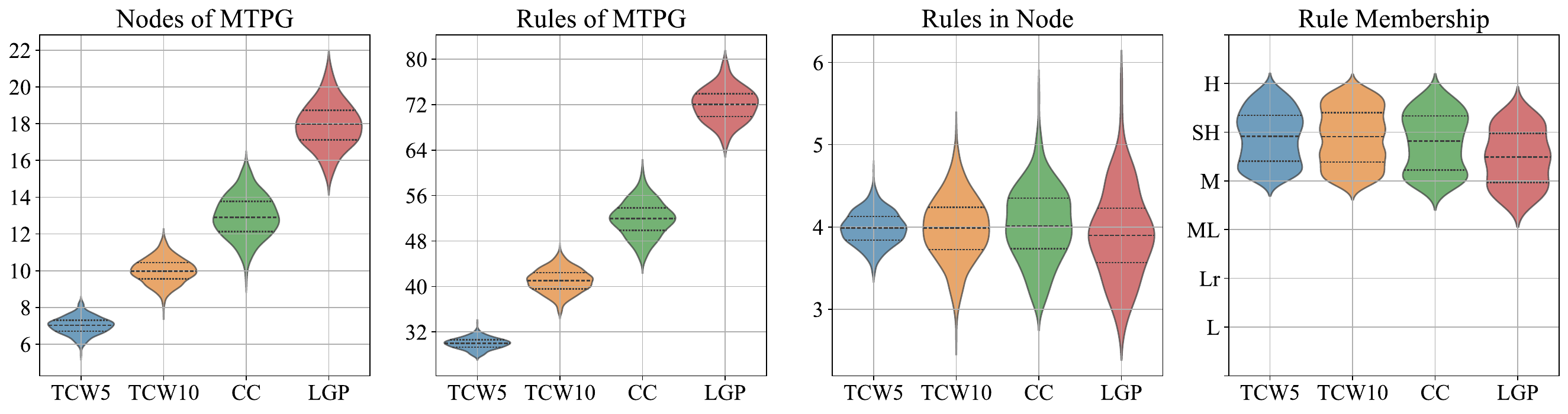}}
  \caption{ Distribution of nodes and rules in MTPG for each task using GPT-4.}
  \label{fig:distribution}
\end{figure*}

\begin{figure}[t]
  \centerline{\includegraphics[width=0.8\linewidth]{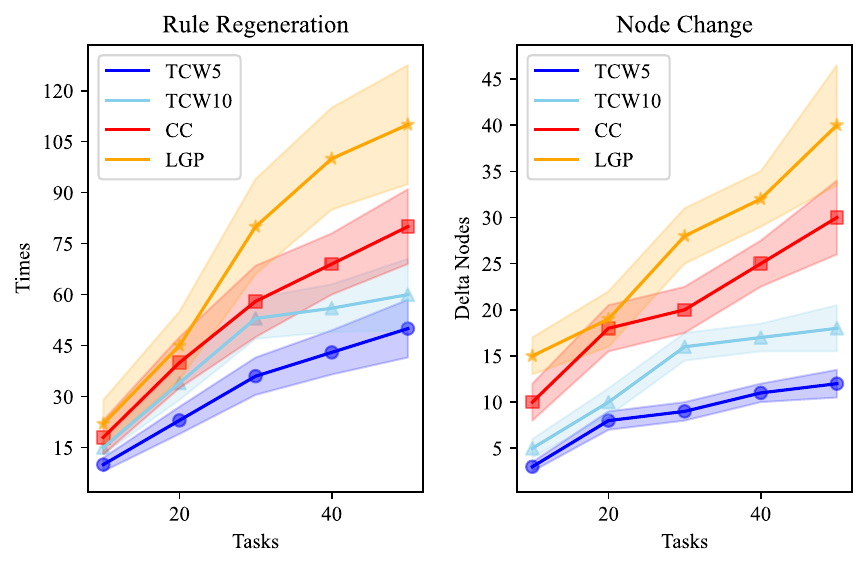}}
  \caption{ Distribution of rule regeneration and node change in MTPG for each Task using GPT-4}
  \label{fig:regeneration}
\end{figure}

\subsection{MTPG's Node and Rule Analysis}
For each task, PA automatically generates a MTPG. To analyze the structure of MTPGs, we conducted a statistical study on the distribution of nodes and rules. As shown in Fig. \ref{fig:distribution}, the average number of nodes increases across datasets in the order: TCW5 $<$ TCW10 $<$ CC $<$ LGP. This reflects the growing task complexity and uncertainty. The tasks of TCW10 include more complex questions than TCW5. CC combines knowledge-logic-mixed questions in every task, requiring more granular decomposition. LGP focuses entirely on logic tasks, which demand the most detailed breakdown. Consequently, MTPGs for LGP have the highest node count. A similar trend is observed in the rule distribution of MTPG. Additionally, the domain-specific rules consistently exhibit membership degrees in the \emph{\{H, SH, M\}} range, confirming both semantic precision and domain relevance to the task.

Fig. \ref{fig:regeneration} presents the changes in the number of subtask rule regenerations and path node variations in the MTPG as the number of tasks increases from 10 to 50. With the expansion of task scale, the frequency of subtask reprocessing rises, leading to an increase in both newly generated rules and node changes. The dynamic trends of rule and node distribution across datasets are consistent with the previously observed static patterns, with the LGP dataset exhibiting the highest number of rule regenerations and the greatest node variation. 

In addition, \emph{Appendix F} presents a post-hoc interpretability analysis of XAgents: when the membership degree of an input sample is \emph{High}, its SHAP values are concentrated above 0.1; when the degree is \emph{Low}, the SHAP values cluster near 0.0. This reveals a strong correlation between XAgents’ input samples and their corresponding predictions, underscoring the model’s outstanding post-hoc interpretability.

Overall, the results indicate that XAgents is capable of performing adaptive task decomposition, akin to human problem-solving in complex or uncertain tasks. 

\subsection{Autonomous Email Reply Case Study}

\begin{figure*}[t]
  \includegraphics[width=01.0\linewidth]{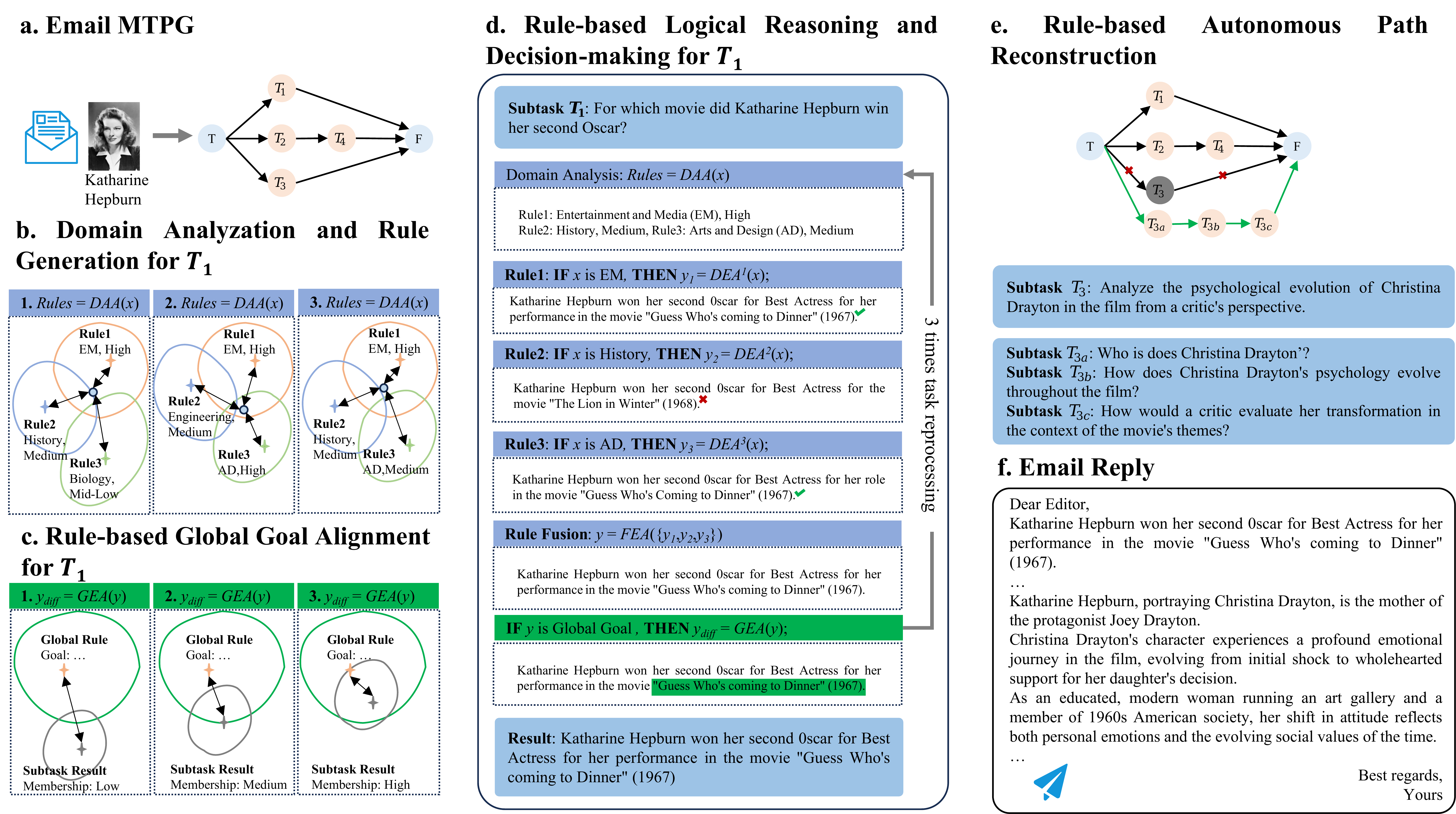}
  \caption{Case study: XAgents applied to a real-world email processing task.}
  \label{fig:email}
\end{figure*}

To further analyze the capabilities of XAgents, we applied it to real-world email processing tasks, as shown in Fig. \ref{fig:email}. This case involves replying to an email from the editor-in-chief of a film review magazine. The processing flow is as follow:
\begin{itemize}
    \item Step 1. PA initializes an MTPG. MTPG includes an original task node, four subtask nodes, and a Fusion Node, as illustrated in Fig. \ref{fig:email} a.
    \item Step 2. For each subtask, ITRDM uses DAA to initialize three domain rules. As shown in Fig. \ref{fig:email} b, DAA analyses the domains of the $T_1$ subtask.
    \item Step 3. Based on the domain rules, the subtasks are processed from different domain perspectives, and all the results are fused by FEA in an adversarial manner to generate a unified output. This output is then matched with the global objective under the global rule.  As shown in Fig. \ref{fig:email} d, ITRDM conducts rule-based reasoning and decision-making for the subtask $T_1$.
    \item Step 4. If it does not match, the process returns to Step 2. If a subtask fails to execute, a task path reconstruction is triggered. As shown in Fig. \ref{fig:email} e, the subtask $T_3$ failed, ITRDM reconstructs the task path by adding new subtasks \{$T_{3a}, T_{3b}, T_{3c}$\}.
    \item Step 5. After all subtasks are finalized, all outputs are fused at the fusion node of the MTPG to form a complete email as shown in Fig \ref{fig:email} f.
\end{itemize}

Based on this case, we analyze the rationale of XAgents from the following perspectives: (i) Global Rule Guiding Cooperation, (ii) Rule-based Semantic Confrontation.

\subsubsection{Global Rule Guiding Cooperation}
\label{subsubsec:global rule guiding}

After PA's analysis, XAgents can determine that the global goal of the email is the discussion of \emph{Katharine Hepburn’s experience and acting skills as an actress}. 

MTPG of the email contains a set of subtasks $\{T, T_{1}, T_{2}, T_{3}, T_{4}, F\}$, as shown in Fig. \ref{fig:email}.a. In MTPG, the subtask $T_1$ is \emph{For which movie did Katharine Hepburn win her second Oscar?}

After analyzing the domain of $T_1$, DAA generated 3 domain rules corresponding to Entertainment and Media (EM), History, and Biology. The respective domain membership degrees are \emph{H, M, and ML}, as illustrated in Fig. \ref{fig:email} b. Following rule-based reasoning, the initial membership degree according to the global goal is \emph{Low}, as shown in Fig. \ref{fig:email} c. The global rule subsequently feeds the deviation from the global goal back to DAA for further adjustment.

The final result, obtained after three reprocessing iterations, achieved strong alignment with the global goal and a high membership degree. XAgents promotes agent cooperation by leveraging global rules to ensure consistency with the overarching objective. 

MTPG’s $T_3$ subtask is \emph{Analyze the psychological evolution of Christina Drayton in the film from a critic's perspective}. This subtask requires multi-domain knowledge and a comprehensive understanding of the entire film, making it a complex issue. Therefore, PA further decomposes $T_3$ into a simpler set of subtasks $\{T_{3a}, T_{3b}, T_{3c}\}$, as shown in Fig. \ref{fig:email} e. This demonstrates that XAgents possess the ability to autonomously reconstruct paths for task processing. Fig. \ref{fig:email} f depicts the $F$ node of MTPG, where the results from all adjacent nodes are fused into a replying email. 


\subsubsection{Rule-based Semantic Confrontation}
\label{subsubsec:semantic confrontation}
During subtask execution, XAgents adopts a rule-based logical reasoning and decision-making framework, enhanced through semantic confrontation to ensure contextually relevant and goal-aligned responses. In Fig. \ref{fig:email}.d, Rule2’s result semantically conflicts with the other results, and the main difference is the movie name. From the semantic analysis, both Rule1 and Rule3 support that the name is \emph{Guess Who's coming to Dinner (1967)} while Rule2 supports that it was \emph{The Lion in Winter (1968)}. 

XAgents solves the semantic conflict and fuses the domain rules’ results into a consistent final result by the two-layer mechanism. (i) We introduce a voting method that assigns a confidence level for the semantics according to the votes. The higher votes, a higher confidence level, while fewer votes result in a lower confidence level. (ii) By considering the domain membership degree, FEA evaluates the confidence level of semantics, awarding a high-confidence level to those semantics that are generated by the domain rule with high membership and a low-confidence level to those with low membership. The relevant prompts are in \emph{Appendix G}. \emph{Appendix H} presents more semantic adversarial cases. By utilizing the two-layer mechanism,  XAgents effectively excludes low-confidence semantics and delivers high-confidence outcomes.

\subsection{Further Tests}



\subsubsection{Ablation Study}
\label{sec:ablation}
\begin{figure}[t]
  \centerline{\includegraphics[width=0.9\linewidth]{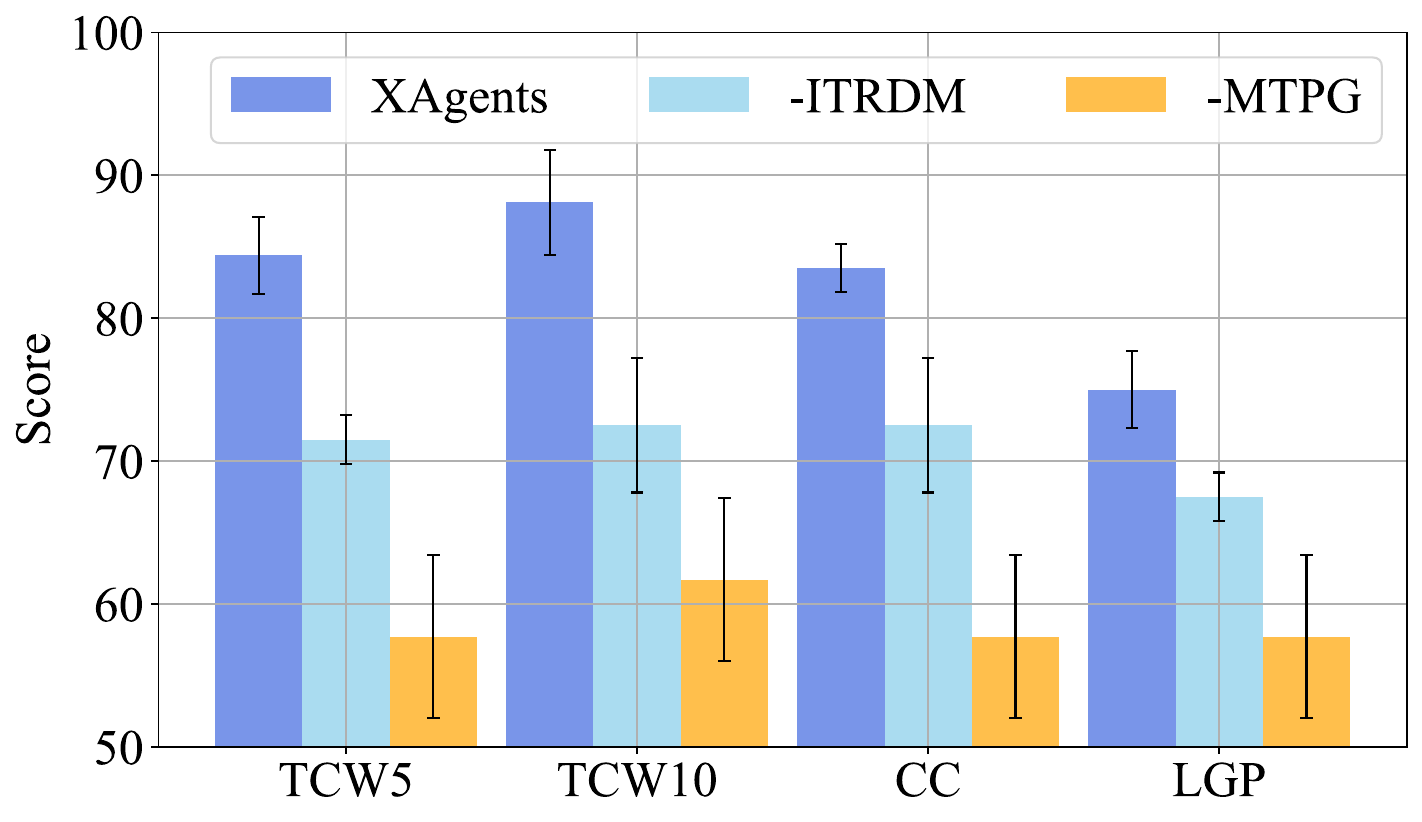}}
  \caption{ Ablation test of XAgents using GPT-4}
  \label{fig:ablation}
\end{figure}

To analyze the contribution of each module to the final outcome, we conducted an ablation study on XAgents, as shown in Fig. \ref{fig:ablation}. We evaluated the performance impact of removing two key components: (i) ITRDM (-ITRDM) and (ii) MTPG (-MTPG). On average, their performance dropped by 11\% and 16\%, respectively. The results confirm that both MTPG and ITRDM play critical roles in enhancing the performance of XAgents.

\subsubsection{Significance Test}
\label{sec:significance}
To further assess statistical significance, we conducted significance tests on XAgents and the baselines across three datasets. Statistical significance was tested for the results of all methods using the Friedman Test method \cite{pereira_overview_2015}. Table \ref{tab:significance} shows that all p-values are less than the significance level ($\alpha$=0.05), indicating statistically significant differences.

\begin{table}
    \setlength{\tabcolsep}{8.5mm}
    \begin{tabular}{lcc}
        \toprule[2pt]
        \textbf{Dataset} &\textbf{statistic}& \textbf{p-value} \\
        \midrule[1pt]
        TCW5&	12.561&	 	0.005\\
        TCW10&	10.920&	 	0.012\\
        CC&    16.039&   	0.007\\
        LGP&	14.039&	 	0.003\\
        \bottomrule[2pt]
    \end{tabular}
     \caption{\label{tab:significance} Significance test of XAgents and baselines using GPT-4.}
\end{table}

\subsubsection{Computational Complexity Test}
\label{sec:complexity scalability}
To evaluate the time and space complexity of XAgents, we measured its runtime, memory consumption, CPU utilization, and API token cost on each task of CC. XAgents reduces memory usage by 44.5\% and token consumption by 28.8\% compared to AgentNet. As shown in Table \ref{tab:complexity}, XAgents demonstrates lower time and space complexity compared to all other multi-agent methods.

\begin{table}
    \setlength{\tabcolsep}{2.2mm}
    \begin{tabular}{lccccc}
        \toprule[2pt]
        \textbf{Methods} &\textbf{Runtime} &\textbf{Memory} &\textbf{CPU}  &\textbf{Tokens}\\
        \midrule[1pt]
        Standard&	10.3s&	1.0MB&	0.01\%  &515\\
        CoT&	    16.0s&	3.0MB&  0.01\%  &819\\
        Self-Refine&24.3s&  3.1MB&	0.01\%  &1,263\\
        SPP&	    125.6s&	26.3MB&	0.21\%  &6,272\\
        AutoAgents&	143.2s&	26.4MB&	0.41\%  &7,836\\
        TDAG&       154.1s& 54.8MB& 0.87\%  &6,900\\
        AgentNet&   136.1s& 44.7MB& 1.01\% &8,451\\
        XAgents&	120.4s&	24.8MB&	0.18\%  &6,010\\
        \bottomrule[2pt]
    \end{tabular}
    \caption{\label{tab:complexity} Computational complexity of XAgents on CC using GPT-4.}
\end{table}

\section{Conclusions}
\label{sec:conclusion}

XAgents introduces a novel mechanism of semantic confrontation between different rules, enabling it to detect and correct logical errors while effectively suppressing hallucinations commonly observed in LLM-generated outputs. 
Compared to existing mainstream multi-agent frameworks, XAgents demonstrates superior capability in handling task uncertainty and in constraining and guiding agent behaviors through explicit rule structures.
Future works will focus on (i) enhanced Rule-Based Explainability, (ii) Rule-Based Systems for Mitigating LLM Hallucinations, and (iii) Optimization of Multipolar Task Graph Structures to improve coordination.
\bibliography{aaai2026}

\end{document}